\pdfoutput=1
\documentclass{article}

 \usepackage[preprint,nonatbib]{neurips_2021}

\usepackage[utf8]{inputenc} 
\usepackage[T1]{fontenc}    
\usepackage{hyperref}       
\usepackage{url}            
\usepackage{booktabs}       
\usepackage{amsfonts}       
\usepackage{nicefrac}       
\usepackage{microtype}      
\usepackage{xcolor}         

\usepackage{graphicx}
\usepackage{lipsum}  
\usepackage{amsmath}
\usepackage[title]{appendix}

\renewcommand{\footnotesize}{\fontsize{8pt}{10pt}\selectfont}

\title{Evolving Evocative 2D Views \\ of Generated 3D Objects}

\author{%
  Eric Chu \\
  MIT Media Lab\\
  \texttt{echu@mit.edu} \\
}

\begin{document}

\maketitle


\begin{abstract}

We present a method for jointly generating 3D models of objects and 2D renders at different viewing angles, with the process guided by ImageNet and CLIP -based models.
Our results indicate that it can generate anamorphic objects, with renders that both evoke the target caption and look visually appealing.






\end{abstract}

\section{Introduction}

While there has been significant effort by both the research and artistic communities to use image models such as BigGAN \cite{brock2018large} and CLIP \cite{radford2021learning} to produce artwork, using deep generative models to create 3D works of art is comparatively underexplored.
In working towards that goal, we are also motivated to model \textit{anamorphic art} \cite{topper2000anamorphosis}, which change or only become recognizable upon certain viewing angles
\footnote{\raggedright Skull when viewed from bottom left: \url{https://en.wikipedia.org/wiki/The_Ambassadors_(Holbein)}}
\footnote{\raggedright Portrait of Ferdinand Cheval in junk heap: \url{https://www.fastcompany.com/3032876/perspective-is-everything-this-anamorphic-sculpture-is-and-isnt-what-it-appears}}
\footnote{\raggedright Ambigram on cover of Gödel, Escher, Bach: \url{https://i.stack.imgur.com/OKBvZ.jpg}}
.
This work also loosely parallels computer vision research aiming for more \textit{interpretable} and \textit{controllable} image generation by jointly learning the 3D generative model and the 2D rendering process \cite{liao2020towards}.






\begin{figure}[!h]
\centering
\includegraphics[width=1.0\linewidth]{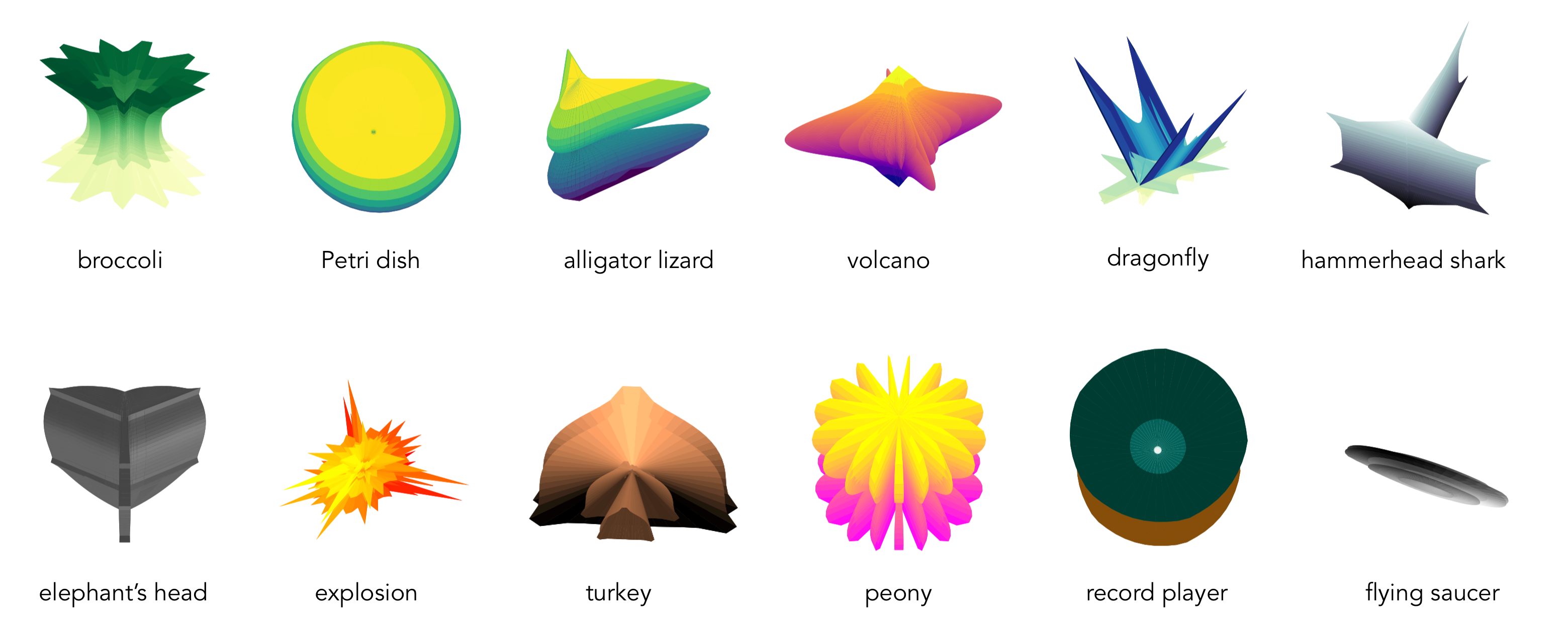}
\caption{
Rendered views of generated 3D objects, under different optimization objectives. Row 1: optimizing for the given class under an ImageNet-based model. Row 2: optimizing for similarity to the text ``A [photo|painting] of ...'' under the CLIP model.
}
\label{fig:results}
\end{figure}

\section{Method and results}

\begin{figure}[!h]
\centering
\includegraphics[width=0.7\linewidth]{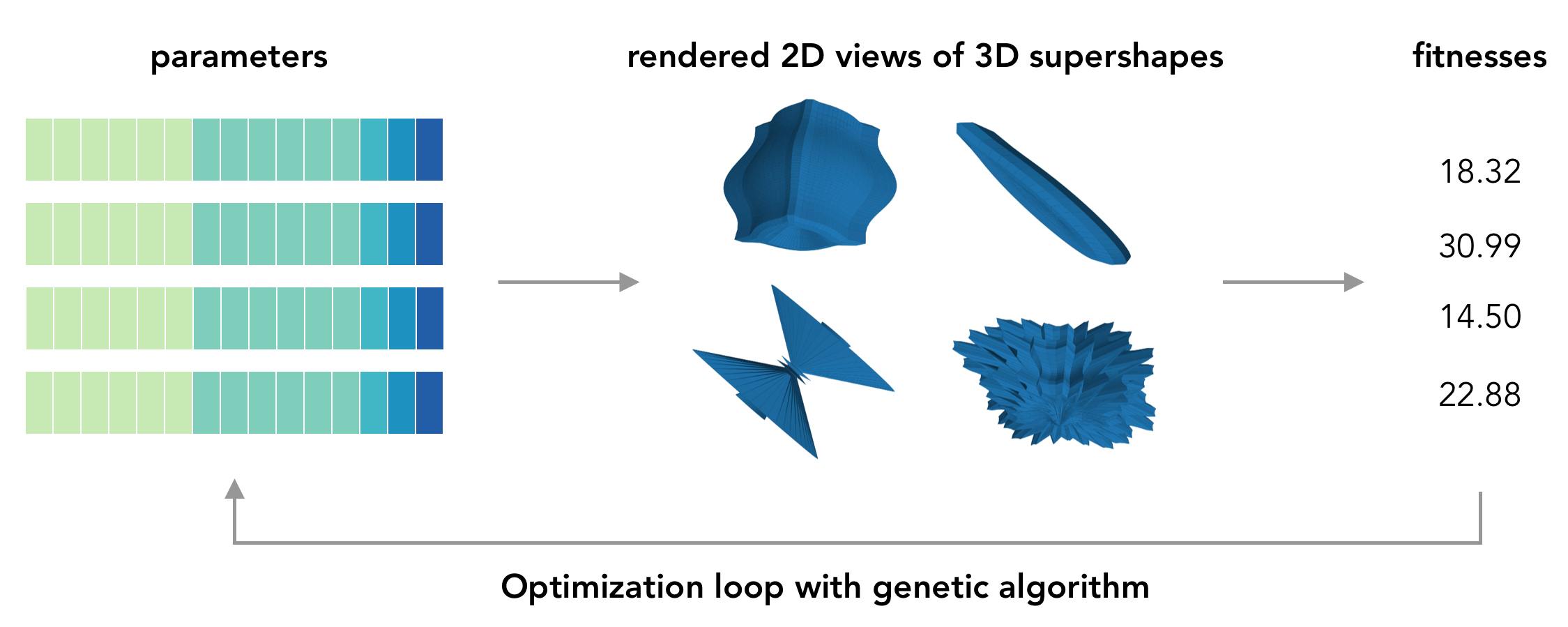}
\caption{Pipeline for generating 3D objects and views.}
\label{fig:pipeline}
\end{figure}

The pipeline for generating 3D structures, and ultimately images of the generated objects, consists of three components: (1) a generator, (2) a scorer, and (3) an optimization loop using a genetic algorithm as shown in Figure \ref{fig:pipeline}.

The \textbf{first component of our generator} is the 3D version of the ``superformula'', a generalization of the superellipse originally introduced as a simple, ``universal'' equation to model a wide range of natural and abstract shapes \cite{gielis2003generic}.
This serves as a powerful and computationally tractable 3D model, but could eventually be replaced by deep generative models \cite{zhou2018voxelnet,nash2020polygen} as they improve.
The 2D equation, with 6 parameters $m,a,b,n_1,n_2,n_3$, is given as:

{\tiny
\begin{align*}
    r(\varphi)=\left(\left|\frac{\cos \left(\frac{m \varphi}{4}\right)}{a}\right|^{n_{2}}+\left|\frac{\sin \left(\frac{m \varphi}{4}\right)}{b}\right|^{n_{3}}\right)^{-\frac{1}{n_{1}}}
\end{align*}
}%

This can be generalized to three dimensions using two instances of the superformula $r_1$ and $r_2$ by:
{\tiny
\begin{align*}
&x=r_{1}(\theta) \cos \theta \cdot r_{2}(\phi) \cos \phi \\
&y=r_{1}(\theta) \sin \theta \cdot r_{2}(\phi) \cos \phi \\
&z=r_{2}(\phi) \sin \phi
\end{align*}
}%

In order to render the 3D surface to a 2D image, the \textbf{second component of our generator} controls the viewing angle by setting three additional parameters -- the elevation, azimuth, and rotation. This can introduce considerable variability and asymmetry into the final product.

These images are then assessed by a \textbf{scorer}. Specifically, the fitness of an image is calculated by either (a) the loss against a specific ImageNet class under a trained MobileNetV3 model \cite{howard2019searching}, or (b) the similarity to a caption under the CLIP  model \cite{radford2021learning}.
As the rendering proecss is non-differentiable, we optimize the 15 parameters in a genetic algorithm -based \textbf{optimization loop}.
The parameters are mutated at a rate of 0.1 and a selection rate of 0.5 while using the roulette wheel selection strategy. We keep a population size of 40 at every iteration. 

\begin{figure}[!h]
\centering
\includegraphics[width=0.8\linewidth]{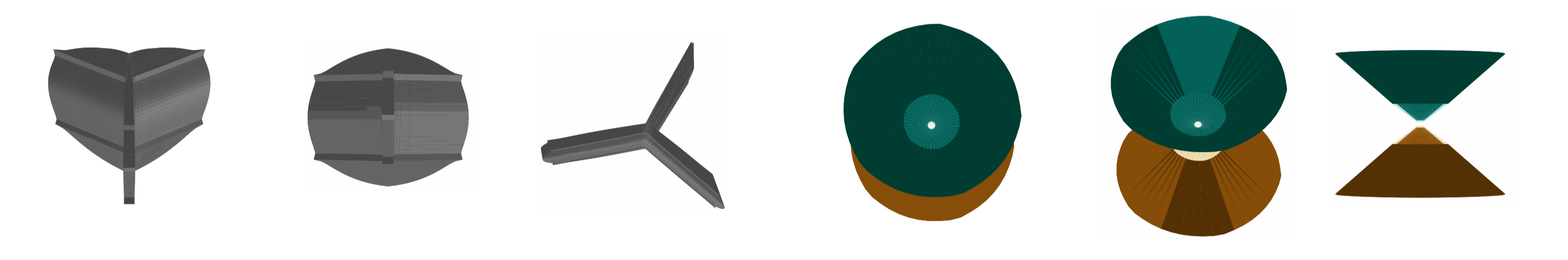}
\caption{
Different views of the generated surfaces for ``elephant's head'' and ``record player''.
}
\label{fig:views}
\end{figure}

As shown in Figures \ref{fig:results} and \ref{fig:views}, our method is able to effectively simulatneously generate surfaces and control the viewing angle in order to evoke the desired object. Often times, the target is only clear at highly specific angles. Finally, we present some examples in the Appendix of (1) explorations of generating views of Richard Serra -style sculptures using a purely 2D-based generative model, and (2) cherry-picked views evolved using novelty search that the author found personally visually appealing as abstract art.


\bibliography{main.bib}
\bibliographystyle{abbrv}









\clearpage

\begin{appendices}



\section{Views of Richard Serra -style sculptures using VQGAN+CLIP}

The artist Richard Serra has a series of sculptures, such as \textit{To Lift}\footnote{To Lift: \url{https://www.moma.org/collection/works/101902}}, that were created by ``applying a verb'' \footnote{Verb list: \url{https://www.moma.org/collection/works/152793?artist_id=5349&page=1&sov_referrer=artist}} on a material such as fiberglass, neon, vulcanized rubber, or lead.

\indent
\textit{``It struck me that instead of thinking what a sculpture is going to be and how you're going to do it compositionally, what if you just enacted those verbs in relation to a material, and didn't worry about the results?''}

We viewed these kind of sculptures as an interesting test bed for material modeling and compositionality (material + verb) for existing image generative models. While one might prefer a joint 3D-2D model as we propose, we investigate the popular VQGAN+CLIP pipeline's ability to replicate and produce new Serra -style art. Shown below in Figure \ref{fig:serra}, we see that the model is able to produce a limited approximation of the target captions.

\begin{figure}[!h]
\centering
\includegraphics[width=1.0\linewidth]{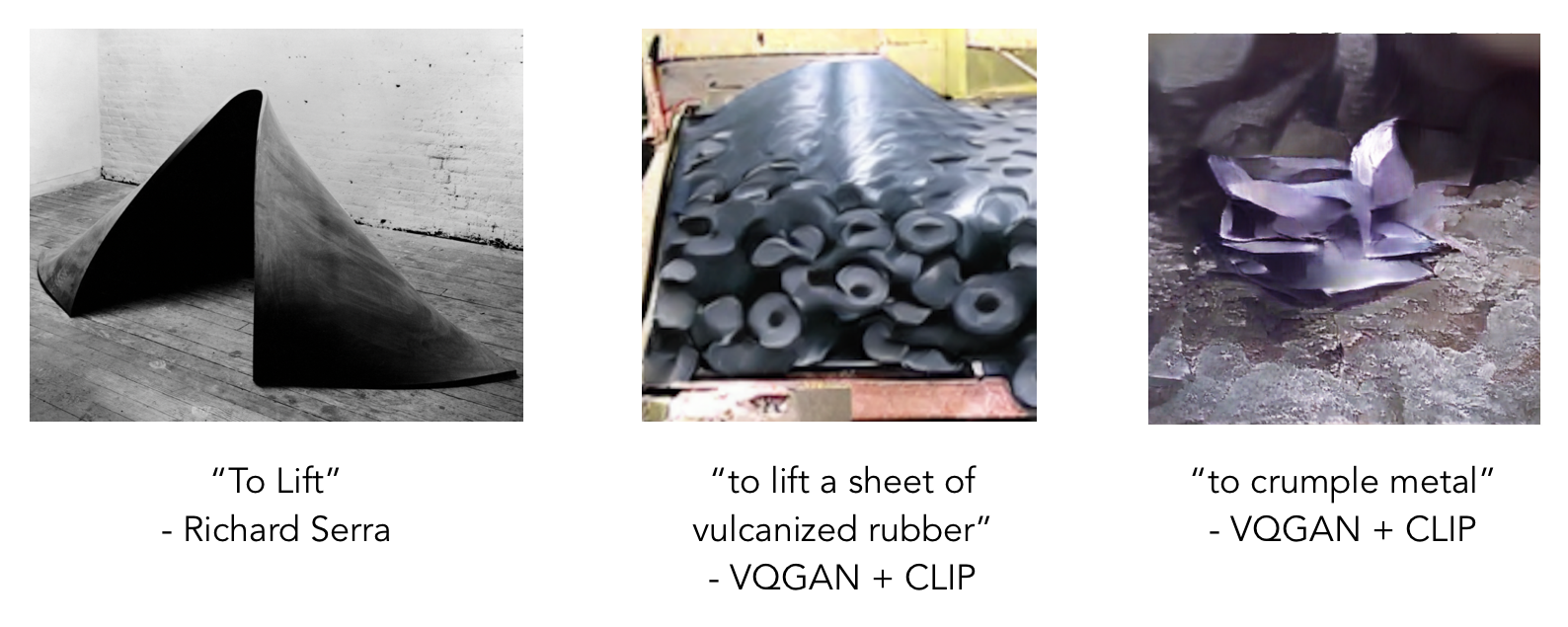}
\caption{
Richard Serra's verb sculpture vs. VQGAN+CLIP generations.
}
\label{fig:serra}
\end{figure}

\section{Abstract art discovered through novelty search}

We also exlored the use of novelty search (instead of optimization against CLIP or an ImageNet model) to generate views. Several examples are shown below:

\begin{figure}[!h]
\centering
\includegraphics[width=0.8\linewidth]{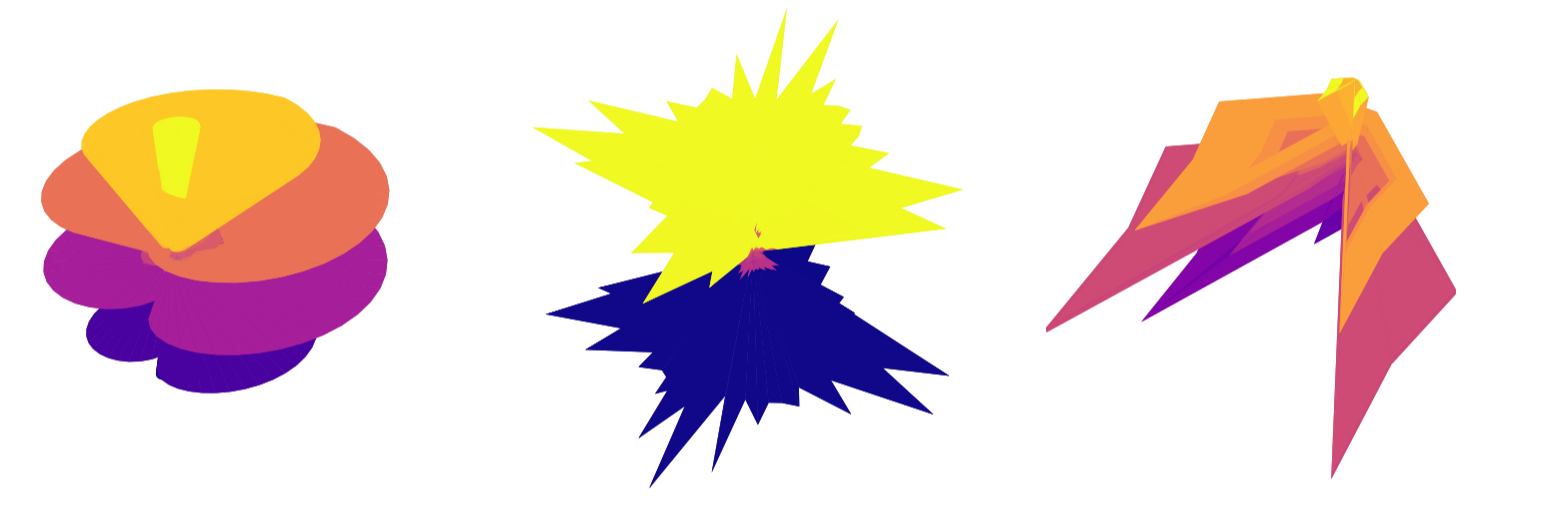}
\caption{
Views discovered during novelty search.
}
\label{fig:novelty}
\end{figure}

\end{appendices}

\end{document}